\documentclass[preprint,review,12pt]{elsarticle}
\usepackage[top=2cm,bottom=2cm,left=1.3cm,right=1.3cm]{geometry}

\usepackage{amssymb}
\usepackage{amsthm}

\usepackage{tabularx,environ}
\usepackage{soul}
\usepackage{caption}
\usepackage{subcaption}
\usepackage{tikz}
\usepackage{amsmath}
\usepackage{multirow}
\usepackage{color}

\newtheorem{prop}{Proposition}
\newtheorem{definition}{Definition}[]

\newtheorem{rem}{Remark}
\usepackage{rotating}
\usepackage{fixmath}
\usepackage{algorithm}
\usepackage[noend]{algpseudocode}
\usepackage{lscape}
\makeatletter
\newcommand{\problemtitle}[1]{\gdef\@problemtitle{#1}}
\newcommand{\probleminput}[1]{\gdef\@probleminput{#1}}
\newcommand{\problemquestion}[1]{\gdef\@problemquestion{#1}}

\NewEnviron{proble}{
  \problemtitle{}\probleminput{}\problemquestion{}
  \BODY
  \par\addvspace{.5\baselineskip}
  \noindent
  \begin{tabularx}{\textwidth}{@{\hspace{\parindent}} l X c}
    \multicolumn{2}{@{\hspace{\parindent}}l}{\@problemtitle} \\
    \textbf{Input:} & \@probleminput \\
    \textbf{Question:} & \@problemquestion
  \end{tabularx}
  \par\addvspace{.5\baselineskip}
}
\newcommand\notsotiny{\@setfontsize\notsotiny{6.3}{6.3}}
\makeatother

\journal{}

\begin{document}

\begin{frontmatter}

\title{Learned upper bounds for the Time-Dependent Travelling Salesman Problem}

\author[salento]{Tommaso Adamo}
\ead{tommaso.adamo@unisalento.it}

\author[salento]{Gianpaolo Ghiani}
\ead{gianpaolo.ghiani@unisalento.it}

\author[salento]{Pierpaolo Greco}
\ead{pierpaolo.greco@unisalento.it}

\author[salento]{Emanuela Guerriero\corref{mycorrespondingauthor}}
\cortext[mycorrespondingauthor]{Corresponding author}

\ead{emanuela.guerriero@unisalento.it}

\address[salento]{Dipartimento di Ingegneria dell'Innovazione,  Universit\`{a} del Salento, \\Via per Monteroni, 73100 Lecce, Italy}

\begin{abstract}
Given a graph whose arc traversal times vary over time, the Time-Dependent Travelling Salesman Problem consists in finding a Hamiltonian tour of least total duration covering the vertices of the graph. The main
goal of this work is to define tight upper bounds for this problem by reusing the information
gained when solving instances with similar features. This is customary in distribution management, where vehicle routes have to be generated over and over again with similar input data. To this aim, we devise an upper bounding technique based on the solution of a classical (and simpler) time-independent Asymmetric Travelling Salesman Problem, where the constant arc costs are suitably defined by the combined use of a Linear Program and a mix of unsupervised and supervised Machine Learning techniques. The effectiveness of this approach has been assessed through a computational campaign on the real travel time functions of two European cities: Paris and London. The overall average gap between our heuristic and the best-known solutions is about 0.001\%. For 31 instances, new best solutions have been obtained.
\end{abstract}

%

\begin{keyword}
time-dependent routing \sep path ranking invariance \sep machine learning \sep travelling salesman problem
\end{keyword}

\end{frontmatter}
\section{Introduction}
\label{intro}
The purpose of this article is to present a Machine Learning (ML) enhanced upper-bound for the \textit{Time-Dependent Travelling Salesman Problem} (TDTSP), defined as follows.  Let  $G:= (V\cup\{0\},A,\tau)$ denote  a time-dependent directed  complete graph, where $V=\{1,\dots,n\}$ is the set of customers, vertex 0 is the depot and $A := \{(i, j) : i \in V, j \in V\}\bigcup\{(0,i):i \in V\}\bigcup\{(i,0):i \in V\}$ is the set of arcs. With each arc $(i, j)\in A$ is associated a travel time function $\tau_{ij}(t)$, representing the travel time of $(i, j)$ if the vehicle leaves node $i$ at time $t$. The TDTSP amounts to determine a least duration tour visiting each customer once, with the vehicle leaving the depot at time 0.\\
\indent In recent years there has been a flourishing of scholarly works in time-dependent routing. See \citep{Gendreau2015} for a review of the field. The contribution \cite{malandraki1992} was the first to address the TDTSP and proposed a \textit{Mixed Integer Programming} (MIP) formulation. An approximate dynamic programming algorithm was devised in \cite{Malandraki199645}, while  two heuristics has been developed in \cite{Li2005}. A simulated annealing heuristic was proposed in \cite{schneider2002} and some metaheuristics were proposed in \cite{Harwood:2013aa}.
Cordeau et al. \cite{Cordeau2014} derived some properties of the TDTSP as well as lower and upper bounding procedures. They also represented the TDTSP as MIP model for which they developed some families of valid inequalities. These
inequalities were then used into a branch-and-cut algorithm that solved instances with up to 40 vertices. 
Arigliano et al. \cite{arigliano2018branch} exploited some properties of the problem and developed a branch-and-bound
algorithm which outperformed the Cordeau et al. \cite{Cordeau2014} branch-and-cut procedure. In \cite{melgarejo2015time} a new global constraint was presented and used in a \textit{Constraint Programming} approach. This algorithm was able to solve instances with up to 30 customers. Recently, Adamo et al. \cite{ADAMO2020104795} proposed a parameterized family of lower bounds, where the parameters are chosen by fitting the traffic data. When embedded into a branch-and-bound procedure, their  lower bounding mechanism allows to solve to optimality a larger number of instances than Arigliano et al. \cite{arigliano2018branch}. Variants of the TDTSP have been examined in \cite{Albiach2008789}, \cite{ARIGLIANO201928}, \cite{montero2017integer} and \cite{Boland2020} (\textit{TDTSP with Time Windows}), in \cite{Helvig2003153} (\textit{Moving-Target TSP}) and in \cite{Montemanni1528314} (\textit{Robust TSP with Interval Data}). Finally, it is worth noting that a scheduling problem, other than the above defined TDTSP, is also known as
Time-Dependent TSP. It amounts to sequence a set of jobs on a single machine in which the processing times depend on the position of the jobs within the schedule (\cite{picard1978}, \cite{fox1980},
\cite{gouveia1995}, \cite{vanderwiel1996},
\cite{miranda2010}, \cite{stecco2008branch}, \cite{godinho2014natural},\cite{abeledo2013time}).\\
\indent In this paper, we propose an upper bounding technique inspired by the new findings of the recent paper \cite{ADAMO2021105446}, where the authors studied a property of time-dependent graphs, dubbed \textit{path ranking invariance}. A time-dependent graph is \textit{path ranking invariant} if the ordering of its paths (w.r.t. travel time) is independent of the start time. The authors showed that, if a graph is path ranking invariant, the solution of a large class of time-dependent vehicle routing problems, including the TDTSP, can be determined by solving suitably defined (and simpler) time-independent routing problems. The authors demonstrated that the ranking invariance property can be checked by solving a (\textit{large}) Linear Programming (LP) problem. If the ranking invariance check fails, they proved that a tight lower bound can be derived from the obtained LP solution.\\
\indent In this paper, we show how the new findings of \cite{ADAMO2021105446} can be further generalized for determining tight upper bounds for the TDTSP. The main idea is to determine a heuristic solution by solving the TDTSP on an \textit{auxiliary} time dependent graph, which satisfies the path ranking invariant property. The travel time functions of the auxiliary graph are determined by generalizing the LP-based approach proposed in  \cite{ADAMO2021105446}. In order to obtain a fast computation of the \textit{auxiliary} travel time functions, we take advantage of the predictive capabilities of a supervised ML technique. Indeed, the ultimate goal is the fast computation of tight upper bounds, 
in those settings in which instances with similar features have to be solved over and over again, as it is customary in distribution management. As stated in \cite{BENGIO2021405}, a \textit{company does not care about solving all possible TSPs, but only theirs}. Therefore, instead of starting every time from scratch in the definition of the auxiliary graph, we insert a learning mechanism in such a way the upper bounding procedure can benefit from previous runs on other instances with similar features. To this aim, we boost our LP-based approach with a mix of supervised and unsupervised techniques in the spirit of \cite{a13120340}. To the best of our knowledge, contribution \cite{a13120340} is the first attempt to use ML to solve a time-dependent routing problem. For a comparative
analysis of machine learning heuristics for solving the classical (time-invariant) Travelling Salesman Problem, see  \cite{Uslan2010ACS}.\\
\indent The paper is organized as follows. 
In Section \ref{sec:2} we provide a problem definition and some background information on the study area.
In Section \ref{sec:ub} we introduce a parameterized family of upper bounds computed by solving the TDTSP on suitably defined \textit{auxiliary} time-dependent graphs. Such family of upper bounds gives rise to an optimization problem aiming to determine the parameter providing the best (minimum) upper bounds. In Section 4 we propose a ML-based heuristic approach for solving such optimization problem. In Section 5 we describe computational experiments on the graphs of two European cities (London and Paris). Finally, we draw some conclusions in Section 6.

\section{Problem definition and backgrounds}\label{sec:2}
Let denote with $[0,T]$ the time interval associated to a single working day.  Without loss of generality we suppose that the travel time functions are constant in the long run, that is $\tau_{ij}(t):=\tau_{ij}(T)$ with $t\geq T$. For the sake of notational convenience, we also use $\tau(i,j,t)$  to designate $\tau_{ij}(t)$. We suppose that  traversal time $\tau_{ij}(t)$ satisfy the \textit{first-in-first-out} (FIFO) property, i.e., leaving the vertex $i$ later implies arriving later at vertex $j$.\\
\indent For any given path $p_k := (i_0, i_1,\dots, i_k)$, the corresponding duration  $z(p_k , t)$ can be computed recursively as:
\begin{equation}\label{z_tau}
z(p_k,t):=z(p_{k-1},t)+\tau_{i_{k-1}i_k}(z(p_{k-1},t)),
\end{equation}
with the initialization $z(p_0,t):=0$. Therefore, a compact formulation of the TDTSP is :
$$\min\limits_{p\in P}z(p,0).$$\\
where $P$ denotes  the set of Hamiltonian tours on the time dependent graph $G:= (V\cup\{0\},A,\tau)$.  
Algorithms developed for the \textit{classical} time-invariant TSP are not able to consider time-varying travel times without essential structural modifications. Nevertheless, we observe that the absence of time constraints implies that time-varying travel times have an impact  on the ranking of solutions of the TDTSP, but they do not pose any difficulty for feasibility check of solutions. A quite natural way of defining a heuristic solution approach is to determine the optimal solution of a  \textit{classical} Asymmetric TSP (ATSP), defined on a graph $G_c=(V\cup\{0\},A,c)$ where $c:A\rightarrow  \mathbb{R}^+$ is a time-invariant (dummy) cost function.  The main issue in this approach is how to determine a time-invariant (dummy) cost function that \textit{mimics} in an effective manner the solutions ranking of the original TDTSP.
In this respect, it can be proved that there always exists a time-invariant (dummy) cost function  such that a least duration route of TDTSP is also a least cost solution of the TSP defined on the time-invariant graph $G_c$, which motivates the following definition.
\begin{definition} [\textbf{Valid cost function}] \label{def1}
A time-invariant cost function $c:A\rightarrow \mathbb{R}^+$ is \textit{valid} for the TDTSP defined on $G=(V\cup\{0\},A,\tau)$, if the least duration solution $p^*=\min\limits_{p\in P}z(p,0)$ corresponds to a least cost solution of the time-invariant ATSP defined on $G_c=(V\cup\{0\},A,c)$, that is:
$$\arg\min\limits_{p\in\mathcal{P}} \sum\limits_{(i,j)\in\mathcal{P}}\tau_{ij}(T)=\arg\min\limits_{p\in\mathcal{P}}z(p).$$
\end{definition}
If we are given a cost function \textit{valid} for an instance of the TDTSP defined on a time-dependent $G=(V\cup\{0\},A,\tau)$, we can determine the least duration solution $p^*$ by exploiting algorithms developed for (\textit{classical}) time invariant ATSP. 
In \cite{ADAMO2021105446} the authors studied the relationship between the concept of \textit{valid} cost function and a property of  time-dependent graphs called \textit{path ranking invariance}.
\begin{definition}[\textbf{Path ranking invariance}]
\textit{A time-dependent graph $G$ is path ranking invariant}, if the following relationship holds for any pair of paths $p'$ and $p''$ of $G$:
$$z(p',t)\geq z(p'',t)\quad \forall t\geq 0.$$
\end{definition}
Since travel time functions are constant in the long run, if a time-dependent graph $G=(V\cup\{0\},A,\tau)$ is path ranking invariant then a \textit{valid} cost function is $c(i,j)=\tau_{ij}(T)$. 

\subsection{The auxiliary graph}
The proposed heuristic algorithm is based on the definition of an auxiliary path ranking invariant graph $\underline{G}=(V\cup\{0\},A,\underline{\tau})$ where each $\underline{\tau}_{ij}(t)$ is an \textit{approximation} of $\tau_{ij}(t)$, with $(i,j)\in A$. Each continuous piecewise linear function $\underline{\tau}_{ij}(t)$ is generated by the travel time model proposed in \cite{Ichoua2003} (IGP model for short), in which each arc $(i,j)\in A$ is characterized by a constant stepwise speed function  $v_{ij}(t)$ and a length $L_{ij}$. We suppose that the horizon is partitioned into $H$ subintervals $[T_h , T_{h+1} ]$ $(h = 0, \dots , H-1)$, with $T_0 = 0$ and $T_H = T$. We assume that all arcs of the auxiliary graph  $\underline{G}$ share a common speed function, such that 
$$v_{ij}(t)=v_h,$$
with $t\in[T_h,T_{h+1}]$, $h = 0, \dots , H -1$ and $(i,j)\in A$.  According to the IGP model, given a start  time $t$ the travel time value $\underline{\tau}_{ij}(t)$ is computed by the following  iterative procedure. 

\begin{algorithm}[H]
\caption{Computing the travel time  $\underline{\tau}_{ij}(t)$}
\label{alg:alg2}

\begin{algorithmic}

\State

\State $k \leftarrow h: t_{h} \le t \le t_{h+1}$

\State $\ell \leftarrow L_{ij};$
\State $t' \leftarrow t + \ell / v_{k};$

\While{$t' > T_{k+1}$}

\State $\ell \leftarrow \ell - v_{k}(T_{k+1}-t);$ 

\State $t \leftarrow T_{k+1};$

\State $t^{\prime} \leftarrow t + \ell/v_{k+1};$

\State $k \leftarrow k+1$

\EndWhile
\Return $t' - t$

\end{algorithmic}

\end{algorithm}

In the IGP model the speed of a vehicle is not a constant over the entire length of arc $(i, j)\in A$ but it changes when the boundary between two consecutive time periods is crossed.
Since the travel speed is a constant stepwise function, the relationship between the  input parameters  and the output value of the IGP model can be expressed in a compact fashion as follows:
\begin{equation}\label{IGP}
L_{ij}=\int_{t}^{t+\underline{\tau}_{ij}(t)}v(\mu)d\mu.
\end{equation}
We denote with $\underline{z}(p_k,t)$ the traversal time of a path $p_k$ at time instant $t$ on the time-dependent graph  $\underline{G}$, that is 
\begin{equation}\label{z_tau_1}
\underline{z}(p_k,t)=\underline{z}(p_{k-1},t )+\underline{\tau}_{i_{k-1}i_k}(\underline{z}(p_{k-1},t)),
\end{equation}
with the initialization $\underline{z}(p_0,t)=0$.  
\begin{prop}\label{G_b_rank_inv}
(Adamo et al. \cite{ADAMO2021105446} ) The time dependent graph $\underline{G}=(V\cup\{0\},A,\underline{\tau})$ is path ranking invariant.
\end{prop}
\begin{proof}
We observe that from (\ref{IGP}) it follows that given a path $p$  we have that:
$$\sum\limits_{(i,j)\in p}L_{ij}=\int_{t}^{t+\underline{z}(p,t)}v(\mu)d\mu,$$
where the notation $(i,j)\in p$ means that the arc $(i,j)\in A$ is traversed by the path $p$. 
This implies that if a path $p'$ is shorter than a path $p''$ then $p'$ is also quicker than $p''$ for any start time $t\in[0,T]$:
$$\sum\limits_{(i,j)\in p'}L_{ij}\leq\sum\limits_{(i,j)\in p''}L_{ij}\Leftrightarrow \underline{z}(p',t)\leq\underline{z}(p'',t), $$ 
which proves the thesis.
\end{proof}

The main implication of Proposition \ref{G_b_rank_inv} is that an upper bound on the TDTSP defined on the \textit{original} graph $G$ can be obtained by solving a \textit{classical} time invariant ATSP with cost coefficients $c(i,j)=\underline{\tau}_{ij}(T)$. Clearly the quality of the obtained upper bound is correlated with the fitting deviation between the original travel time function $\tau$ and its \textit{approximation} $\underline{\tau}$. Minimizing such \textit{fitting deviation}  is the main  idea underlying the family of parameterized upper bounds presented in the following section.

\section{A family of parameterized upper bounds}\label{sec:ub}
In this section we define a family of parameterized upper bounds $\underline{z}_{\Omega}$, where parameters $\Omega$ constitute an ordered set of time instants. Given set $\Omega$, upper bound $\underline{z}_{\Omega}$ is determined by solving the TDTSP  on an auxiliary path ranking invariant graph $\underline{G}_{\Omega}=(V,A, \underline{\tau}_{\Omega})$. The travel time function $\underline{\tau}_\Omega$ is an approximation of the original travel function $\tau$. In particular  $\underline{\tau}_\Omega$ is generated by the IGP model and satisfies relationship (\ref{IGP}). 
We recall that the IGP parameters are: the set of speed breakpoints, the speed values and the length of the arcs. We make use of the given \textit{upper-bound parameter} $\Omega$ to model the set of IGP speed breakpoints, i.e. $\Omega=\{T_0,\dots,T_H\}$, with $H=|\Omega|-1$.  Then speed values and length of arcs are prescribed by a linear program, which aims to minimize \textit{the fitting deviation} between the original $\tau$ and its parameterized approximation $\underline{\tau}_{\Omega}$. The main idea underlying the linear program is that the equalities (\ref{IGP}) imply that the travel time functions $\tau$ and $\underline{\tau}_{\Omega}$ are perfect fit if the following relationship holds for each arc $(i,j)\in A$ and time instant $t\in T$:
\begin{equation}\label{eq:1}
L_{ij}-\int_{t}^{t+\tau_{ij}(t)}v(\mu)d\mu=0.
\end{equation}
The objective function aims to minimize a \textit{fitting deviation} given by the violations of equality constraints (\ref{eq:1}). Due to the continuous time nature of  (\ref{eq:1}), we define a surrogate of the \textit{fitting deviation} by evaluating (\ref{eq:1}) only for time instants belonging to a set $\Omega_{ij}$, that is:
\begin{equation}\label{eq:1_1}
L_{ij}-\int_{T_h}^{T_h+\tau_{ij}(T_h)}v(\mu)d\mu=0,
\end{equation}
with $h=0,\dots,|\Omega_{ij}|-1$ and $(i,j)\in A$. The set $\Omega$ is defined as the union set of $\Omega_{ij}$, with $(i,j)\in A$, i.e. $\Omega=\bigcup\limits_{(i,j)\in A}\Omega_{ij}$.\\
\indent Let define the coefficient $a_{ijkh}$ as follows:
\begin{equation} \nonumber
a_{ijkh} =     \left\{ \begin{array}{cr}
         \min(T_{h+1}-T_{h},\max(0,T_k+\tau_{ij}(T_k)-T_{h}))&
          k\leq h \\ 
         0  & otherwise 
                \end{array}\right.
\end{equation}
with $(i,j)\in A$, $h,k=0,\dots,|\Omega_{ij}|-1$.

Since $v(t)$ is constant stepwise, relationship (\ref{eq:1_1}) can be expressed by the following linear equality:
\begin{equation}\label{eq:2}
\sum\limits_{h=0}^{|\Omega_{ij}|-1}a_{ijkh}\times v_h= L_{ij} +s_{ijk},
\end{equation}
where  the \textit{free-sign} variable $s_{ijk}$ models the \textit{violation} of the right-hand-side of (\ref{eq:1_1}) with respect to $L_{ij}$, with $(i,j)\in A$, $k=0,\dots,|\Omega_{ij}|-1$. The proposed linear program determines a speed function $v(t)$ and the corresponding right-hand-sides of (\ref{eq:2}), which we denote with $x_{ijk}$: since it represents a length we require that $x_{ijk}\geq 0$, with $(i,j)\in A$, $k=0,\dots,|\Omega_{ij}|-1$. 
We model the \textit{maximum fitting deviation} between the original travel time function $\tau(i,j,t)$ and  $\underline{\tau}_{\Omega}(i,j,t)$ as
$$\zeta_{ij}=\max\limits_{k\in[0,\dots,|\Omega_{ij}|-1]}x_{ijk}-\min\limits_{k\in[0,\dots,|\Omega_{ij}|-1]}x_{ijk},$$
with $(i,j)\in A$. Quantity $\zeta_{\Omega}=\sum\limits_{(i,j)\in A}{\zeta_{ij}}$ represents an \textit{approximated} measure of the \textit{total fitting deviation} associated to the auxiliary graph $\underline{G}_{\Omega}$. We determine the auxiliary graph $\underline{G}_{\Omega}$, so that the corresponding travel time function $\underline{\tau}_{\Omega}$ minimizes the value of $\zeta_{\Omega}$.
To this aim, we formulate the following linear program (\ref{obj})-(\ref{c5}), where $\underline{x}_{ij}$ and $\overline{x}_{ij}$ model, respectively, the minimum and maximum value of the  variables $x_{ijk}$, with $(i,j)\in A$ and $k=0,\dots,|\Omega_{ij}|-1$. A solution of such linear programming model represents the parameters of a constant piecewise function $y(t)$ and constant values $x_{ijh}$, with $h=0,\dots,|\Omega_{ij}|-1$ and $(i,j)\in A$. The  continuous variable $y_h$ represents the value of $y(t)$ during the $h-th$ time interval, that is:
$$y(t)=y_h,$$
with $t\in[t_h,t_{h+1}]$ and $h=0,\dots,|\Omega|-1$. 
\begin{equation}\label{obj}
\zeta_{\Omega}^*:=\min \sum\limits_{(i,j)\in A}{\overline{x}_{ij}-\underline{x}_{ij}}
\end{equation}
s.t.
{\allowdisplaybreaks
\begin{flalign}
\label{c1}
&\sum\limits_{h=0}^{|\Omega_{ij}|-1}a_{ijkh}\cdot y_h=x_{ijk}\quad\quad
k=0,\dots,|\Omega_{ij}|-1\quad(i,j)\in A&&\\
\label{c3}
&\underline{x}_{ij}\leq x_{ijk}\quad\quad\quad k=0,\dots,|\Omega_{ij}|-1,  (i,j)\in A&&\\
\label{c4}
&\overline{x}_{ij}\geq x_{ijk}\quad\quad\quad k=0,\dots,|\Omega_{ij}|-1 ,  (i,j)\in A&&\\
\label{c6}
& x_{ijk}\geq0,\quad\quad\quad k=0,\dots,|\Omega_{ij}|-1 ,  (i,j)\in A&&\\
\label{c8}
& \underline{x}_{ij}\geq0 \quad\quad\quad\quad\quad\quad\quad(i,j)\in A &&\\
\label{c10}
& \overline{x}_{ij}\geq0 \quad\quad\quad\quad\quad\quad\quad(i,j)\in A &&\\
\label{c5}
& y_{h}\geq\rho\quad\quad\quad h=0,\dots,|\Omega|-1&&
\end{flalign}
}

Objective function (\ref{obj}) aims to determine a step function $y^*(t)$ that minimizes the total maximum fitting deviation between the original travel time function $\tau$ and its approximation $\underline{\tau}_{\Omega}$. Constraints (\ref{c1}) state the relationship between $y(t)$ and $x$ variables. 
Constraints (\ref{c3}) and  (\ref{c4}) model the relationship between $\underline{x}_{ij}$, $\overline{x}_{ij}$ and continuous variables $x_{ijk}$. Constraints (\ref{c6}), (\ref{c8}), (\ref{c10}) and (\ref{c5}) describe the non-negative conditions on the decision variables. In particular, in order to  cut off the trivial (pointless) solution $y(t) = 0$ for $t \geq0$, constraints (\ref{c5}) state that the constant stepwise linear function $y(t)$ has to be greater  or equal than the input parameter $\rho>0$.\\
Let $y^*(t)$ and $x^*$ denote, respectively, the step function and the $x$ values associated with the optimal solution of the linear program (\ref{obj})-(\ref{c5}).
Moreover, we denote with $\tilde{x}^*_{ij}$ the average of the $x$ values associated to arc $(i,j)\in A$ in the optimal solution, that is:
$$\tilde{x}^*_{ij}=\sum\limits_{h=0}^{|\Omega_{ij}|-1}\frac{x^*_{ijh}}{|\Omega_{ij}|}.$$ 
We observe that the linear program does not \textit{directly} prescribe the IGP parameter $L_{ij}$, with $(i,j)\in A$. Indeed, according to (\ref{eq:2}) we have that:
$$x^*_{ijk}=L_{ij}+s_{ijk},$$
 where, we recall, $s_{ijk}$ quantifies the violation of equality (\ref{eq:1_1}), with $(i,j)\in A$ and $k=0,\dots,|\Omega_{ij}|-1.$
 Since $L_{ij}$ denotes the IGP length associated with $\underline{\tau}_{\Omega}$, from (\ref{eq:2}) we have that
$$\int_{t_k}^{t_k+\tau(i,j,t_k)}v(\mu)d\mu-\int_{t_k}^{t_k+\underline{\tau}_{\Omega}(i,j,t_k)}v(\mu)d\mu=s_{ijk},$$
that is the lower the \textit{absolute} value of equality (\ref{eq:1_1}) \textit{violation} (i.e. $|s_{ijk}|$), the lower the \textit{absolute error} made by approximating $\tau(i,j,t_k)$ with $\underline{\tau}_{\Omega}(i,j,t_k)$, with $t_k\in\Omega_{ij}$ and $(i,j)\in A$. Since $\tilde{x}^*_{ij}$ minimizes the mean squared \textit{violation} of equality (\ref{eq:1_1}), i.e. 
$$\tilde{x}^*_{ij}=\arg\min\limits_{L_{ij}}\sum\limits_{k=0}^{|\Omega_{ij}|-1}\frac{(x^*_{ijk}-L_{ij})^2}{|\Omega_{ij}|},$$
we (\textit{heuristically}) minimize such travel time \textit{approximation} errors by generating the travel time function $\underline{\tau}_{\Omega}(i,j,t)$ with the following IGP input parameters:
 $$v(t)=y^*(t), \quad L_{ij}=\tilde{x}^*_{ij},$$
 with $(i,j)\in A$.
Finally, we recall that the travel time function $\underline{\tau}_{\Omega}(i,j,t)$ satisfies relationship (\ref{IGP}), and, therefore, the auxiliary graph is path ranking invariant.
Summing up, given a set of time instants $\Omega=\bigcup\limits_{(i,j)\in A}\Omega_{ij}$ and a time dependent graph $G$, the proposed upper bounding procedure consists of  three main steps. 
\begin{itemize}
\item \textbf{STEP 1}. Solve linear program (\ref{obj})-(\ref{c5}). Set the travel speed function $v(t)$ equal to $y^*(t)$. Similarly set $L_{ij}$ to  $\tilde{x}^*_{ij}$ for each $(i,j)\in A$. 
\item \textbf{STEP 2}. Determine solution $\underline{p}_{\Omega}^*$ as the least cost solution of the following time-independent ATSP:
$$\min\limits_{p\in \mathcal{P}}\sum\limits_{(i,j)\in p}\underline{\tau}_{\Omega}(i,j,T).$$
\item  \textbf{STEP 3}. Compute upper bound $\underline{z}_{\Omega}$ by evaluating $\underline{p}_{\Omega}^*$ w.r.t. the original travel time function $\tau$ that is:
$$\underline{z}_\Omega=z(\underline{p}_{\Omega}^*,0)$$
\end{itemize}

\indent We finally observe that in order to find the least upper bound, the following optimization problem has to be solved:

\begin{equation}\label{c9}
\min\limits_{\Omega}\underline{z}_{\Omega},
\end{equation}
where $\underline{z}_{\Omega}$ is evaluated according to the proposed three-steps procedure. 
A \textit{simple} heuristic for solving (\ref{c9}) is to set each $\Omega_{ij}$ equal to a discretization $\mathcal{D}$ of the planning horizon. In this case we refer to the three-steps procedure computing the upper bound $\underline{z}_{\mathcal{D}}$ as PL-enhanced heuristic (PL-HTSP for short). The main drawback of the PL-HTSP heuristic is that the computation of a tight upper bound value $\underline{z}_{\mathcal{D}}$ might require the solution of a \textit{large} Linear Program.
In the following section we devise a machine learning based heuristic for solving (\ref{c9}) aiming to overcome this drawback. In particular, we exploit the predictive capabilities of machine learning in order to carefully select $\Omega$ as a (quite small) subset of time instants in $\mathcal{D}$.   In this case, we refer to the three-steps upper bounding procedure computing $\underline{z}_{\Omega}$ as MLPL-enhanced heuristic (MLPL-HTSP for short).

\section{Learning to enhance upper bounds}
\label{sec:approach}
In this section, we propose a learning mechanism for determining set $\Omega$. Then upper bound $\underline{z}_{\Omega}$ is computed according to the three-steps upper bounding procedure illustrated in the previous section. As stated in Section 1, the goal is to determine "good" upper bounds, 
in those settings in which instances with similar features have to be solved over and over again, as it is customary in distribution management. Instead of starting every time from scratch in the definition of the auxiliary graph $\underline{G}_{\Omega}$, we devise a learning mechanism so that our upper bounding procedure can benefit from previous runs on other instances with similar features.\\
\indent The idea of \textit{bounds based on an auxiliary path ranking invariant graphs} is inspired by \cite{ADAMO2021105446}, where the authors devised a sufficient  condition for determining the optimal solution of (\ref{c9}). They proposed an iterative procedure to determine $\Omega$. Then they considered a \textit{minimax} variant of  (\ref{obj})-(\ref{c5}), where, basically, relationship (\ref{eq:2}) is modeled as a  \textit{lower approximation}, i.e. $s_{ijk}\geq0$ with $(i,j)\in A$ and $k=0,\dots, |\Omega_{ij}|-1$. The authors  proved that if $\zeta_{\Omega}^*=0$, then  $\underline{\tau}_\Omega$ and $\tau$ are perfect fit and, therefore,  $\underline{p}_{\Omega}^*$  is a tour of  least total duration on $G=(V\bigcup \{0\},A,\tau)$.  If the  optimality check fails, i.e. $\zeta_{\Omega}^*>0$, they proved that $\underline{\tau}_{\Omega}$ is a lower approximation of $\tau$ and the total duration of $\underline{p}_{\Omega}^*$ on the \textit{less congested} graph $\underline{G}_{\Omega}$ is a lower bound for the optimal solution of TDTSP defined on the original graph $G$. As stated by the authors in \cite{ADAMO2021105446}, their upper bound was a by-product of the search for "good" TDTSP lower bounds.\\
In this research work, we aim to devise a machine learning approach where the main goal is to enhance the upper bound $\underline{z}_{\Omega}$. In particular we propose a mechanism for  \textit{learning} the relationship between set $\Omega$ and the optimal solutions of the TDTSP defined on the original time-dependent graph $G$. 
We start by observing that there exists a finite and discrete set $\Omega^*$, consisting of all (\textit{feasible}) arrival times:  if $t$ belongs to $\Omega^*$, then there exists on $G$ a feasible tour $p\in P$ with $t$ corresponding to the arrival time at a node $i\in V$. That such set $\Omega^*$ exists is based on the observation that there is a finite number of feasible tours.  
\begin{rem}\label{rem:1} If $\zeta_{\Omega^*}^*=0$, then for each arc $(i,j)\in A$ and time instants $t\in \Omega^*$, we have that:
$$\underline{\tau}_{\Omega^*}(i,j,t)=\tau(i,j,t)$$
and therefore, upper bound $\underline{z}_{\Omega^*}$ is optimal, that is $\underline{z}_{\Omega^*}=\min\limits_{p\in P}z(p,0).$
\end{rem}
The main limit of the sufficient optimality condition stated in Remark \ref{rem:1} is that determining the entire $\Omega^*$ is computationally challenging. To overcome this drawback, we take advantage  of the predictive capabilities of supervised ML techniques, in order to determine a set $\Omega$ such that the arrival times associated to optimal solutions have a good chance of being included in  $\Omega$. We denote with $f_i$ a prediction (obtained through a supervised ML method) of the \textit{expected time of arrival} (ETA) at customer $i$ in an optimal solution. We observe that the ranking among arcs might deeply changes during the planning horizon on the original graph $G$. On the other hand, the path ranking invariance of the auxiliary graph $\underline{G}_{\Omega}$ implies also an \textit{arc} ranking invariance. The intuition is that, by taking a snapshot \textit{around} the optimal arrival times (of similar instances  previously solved), we have a good chance of embedding in the auxiliary graph $\underline{G}_{\Omega}$ the arc ranking associated to the set of quickest tours of the original graph. For this purpose, we require that the maximum fitting deviation  between the original travel time function $\tau(i,j,t)$ and  $\underline{\tau}_{\Omega}(i,j,t)$ is minimized for each arc $(i,j)\in A$ in the time interval $[f_i-\epsilon_i,f_i+\epsilon_i]$, where $\epsilon_i>0$ represents the mean absolute error associated to $f_i$, with $i\in V$. \\
In particular, we first define a discretization $\mathcal{D}$ of the time horizon. Then for each node $i$ we select the subset $S_i$ of $\mathcal{D}$ as follows:
$$S_i=\{t \in [f_i-\epsilon,f_i+\epsilon]\wedge t \in \mathcal{D}\}$$
In the definition of the approximation travel time $\underline{\tau}_{\Omega}$, all arcs $(i,j)\in A$ outgoing the node $i\in V$ share a common set $\Omega_{ij}$ corresponding to the set $S_i$, i.e. $\Omega_{ij}= S_i$. Therefore in the MLPL-HTSP, the travel time $\underline{\tau}_{\Omega}$ is determined by solving the linear program (\ref{obj})-(\ref{c5}), where the role of $\Omega_{ij}$ is played by the subset $S_i$ in the constraints (\ref{c1})-(\ref{c6}), with $i=1,\dots,n$.   

\subsection{ETA estimation}
In order to estimate the ETA of a customer $i$ in an optimal solution, an artificial neural network (ANN) is used in conjunction with an exact algorithm for the TDTSP \cite{arigliano2018branch}. The chosen ANN  is a \textit{Multilayer Perceptron Regressor} (MPR) \cite{aggarwal2018neural}, consisting of at least three layers of nodes: an input layer, one or more hidden layer and an output layer. Except for the input nodes, each node uses a nonlinear activation function. Firstly customers are aggregated and the service territory is divided into a number of zones $K$. The customer aggregation is an unsupervised learning technique that aims to partition the customers of the training set into $K$ clusters of equal variance, where  the sum of intra-cluster Euclidean distances is minimized. In our experimentation, we used a $K$-means algorithm \cite{macqueen1967some}.  For each training instance, an average zone ETA, named $ZETA_k$, is determined, with $k=1,\dots,K$. In particular, the arrival times at the customers are computed by the exact algorithm of \cite{arigliano2018branch}. The neural network has $K$ inputs and $K$ outputs: the inputs are constituted by  the number $n_k$ of customers in each of the $k$ zones (i.e. the customer distribution in the network); the outputs are the $K$  $ZETA_k$ estimates ($k = 1, \dots, K$). It is worth noting that, if $K$ is large the predictions are expected to be more accurate but the training phase would require a huge number of instances. On the other hand, a small value of $K$ implies a large  variability of the ETA inside a zone, which has a disadvantageous effect on the ETA estimation of individual customers. The optimal number of zones $K$ was determined in a preliminary experimentation.

\section{Computational Experiments}
\label{sec:results}
The quality of the proposed upper bounding procedure was empirically assessed through a computational campaign. The machine learning component of the MLPL-HTSP algorithm was implemented in Python (version 3.6). 
The Multilayer Perceptron Regressor implementation was taken from the \textit{sklearn} neural network library (method MLPRegressor) while the K-means implementation came from the \textit{sklearn} cluster library (K-means method). The training instances were solved to optimality (or near-optimality) using a Java implementation of the branch-and-bound scheme proposed in \cite{arigliano2018branch} enhanced with the lower bound proposed in \cite{ADAMO2021105446}. A time limit of an hour was imposed. The linear program (\ref{obj})-(\ref{c5}) was solved with IBM ILOG CPLEX 12.10. The instances of the Asymmetric TSP have been solved by means of \cite{carpaneto1995exact}. All the codes were tested on a Linux machine clocked at 2.67 GHz and equipped with 8 GB of RAM. 
We considered the instances generated by Adamo et al. \cite{a13120340} and based on the real travel time functions of two major European cities: Paris and London.

\subsection{Parameter tuning}
In a preliminary tuning we have selected the most appropriate combination of parameters.
Our datasets contained approximately $6-700$ instances with 	$50$ customers each: $90\%$ has been assigned to the training set, while the remaining $10\%$ to the test set. The neural network settings providing the best results, in terms of strength of captured relationships were: three layers, hyperbolic tangent activation function, five neurons in the hidden layer, LBFGS solver and constant learning rate.
As far as customer aggregation is concerned, Table \ref{table-1} and Table \ref{table-2} reports the neural network mean errors (in minutes) for each zone.
For London, $8$ clusters gave the best results in terms of coefficient of determination ($R^2$), whilst for Paris $6$ zones were the best case for neural network performance. It is worth noting that the $R^2$ scores ($= 0.53$ for the London instances and $= 0.60$ for the Paris instances) suggest a moderate effect size. We set parameter $\epsilon_i$ equal to the mean absolute error of the zone, which the customer $i\in V$ belongs to. We considered a 5-minutes time unit for the  discretization $\mathcal{D}$ of the planning horizon. Finally, we set $\rho$ equal to $1/ \min\limits_{h=0,\dots,|\Omega|-1}(T_{h+1}-T_h)$.

\begin{table}[ht]
\centering
\scriptsize
\caption{Mean errors in the London instances \label{table-1}}
\begin{tabular}{|c|c|c|c|}
\hline
\textbf{Zone} & \textbf{Mean error} & \textbf{Mean absolute error} & \textbf{Standard error} \\\hline
1            & 7.68       & 36.78               & 55.16                   \\
2            & -4.61      & 29.23               & 37.19                   \\
3            & 8.32       & 26.94               & 35.51                   \\
4            & -1.93      & 27.34               & 36.87                   \\
5            & -2.68      & 28.78               & 46.21                   \\
6            & 8.69       & 56.68               & 69.21                   \\
7            & 2.54       & 24.60                & 32.31                   \\
8            & 6.68       & 54.00                  & 64.84                   \\\hline
\textbf{Average}  & 3.09       & 35.54               & 47.16   \\ \hline              
\end{tabular}
\end{table}

\begin{table}[ht]
\centering
\scriptsize
\caption{Mean errors in the Paris instances\label{table-2}}
\begin{tabular}{|c|c|c|c|}
\hline
\textbf{Zone} & \textbf{Mean error} & \textbf{Mean absolute error} & \textbf{Standard error} \\\hline
1            & -1.02      & 18.55               & 23.74                   \\
2            & 2.40        & 15.29               & 20.14                   \\
3            & 0.74       & 19.69               & 24.30                    \\
4            & -2.78      & 28.85               & 36.53                   \\
5            & 5.53       & 44.65               & 52.49                   \\
6            & 1.33       & 24.00                  & 29.55                   \\\hline
\textbf{Average}  & 1.03       & 25.17               & 31.13    \\\hline              
\end{tabular}
\end{table}

\subsection{Computational results}
As illustrated in the previous section, the predictive capabilities of the ML-techniques have been exploited for the fast computation of two $\Omega$ sets, associated to London and Paris respectively.  Then the two testsets were solved by the MLPL-HTSP algorithm.
The computational results  are presented in Tables \ref{results-london} - Table \ref{results-paris}, under the following headings:
\begin{itemize}
\item the name of the test instance,
\item the objective value $BK$ in minutes of the best-known solution determined by the exact algorithm proposed  in \cite{arigliano2018branch} enhanced with the lower bound proposed in \cite{ADAMO2021105446}, with a time limit of 1 hour;
\item the objective value $\underline{z}_{\Omega}$ in minutes of the MLPL-HTSP solution;
\item the percentage of improvement $DEV$ of $\underline{z}_{\Omega}$ with respect to $BK$, computed as:
$$DEV=\frac{\underline{z}_{\Omega}-BK}{BK};$$
\item $Time$ in seconds spent to determine $\underline{z}_{\Omega}$.
\end{itemize}
If $\underline{z}_{\Omega}$  is a new best-known solution, it is indicated in bold.
The average running times are 18.28 seconds for London instances and 12.46 seconds for Paris instances.
The average percentage deviation between MLPL-HTSP result and the best-known solution is $0.23\%$ for London instances and $-0.18\%$ for Paris instances. 
In the worst case, the percentage deviation is $2.15\%$ and in 31 cases a new best-known solution is obtained. For 38 instances, the MLPL-HTSP heuristic also obtains the best known solution, whilst for 100 out of 140 instances the absolute value $|BK-\underline{z}_{\Omega}|$ is less or equal than 1 minute, which is the smallest time unit normally considered in real vehicle routing problems inside large cities.\\
\indent We  have also examined the impact of both the linear program (\ref{obj})-(\ref{c5}) and the machine learning algorithm. For this purpose  we have implemented a baseline heuristic HTSP, where the auxiliary graph $\underline{G}$ is time-independent, with the constant value associated to each arc $(i,j)\in A$ set equal to  $\max\limits_{t\in[0,T]}\tau_{ij}$, for each $(i,j)\in A$. Table \ref{tab:compsol} and Table \ref{tab:comptime} report results for all three heuristics: column headings are self explanatory. Results associated to the PL-HTSP highlight that the computation of the approximation $\underline{\tau}_{\Omega}$ provides a remarkable increase of both the solution quality and the computing time w.r.t. the baseline heuristic HTSP. 
It is by leveraging the machine learning that the MLPL-HTSP heuristic obtains both solution quality improvement and a reduction (by an order of magnitude) of the computing time w.r.t. the PL-HTSP heuristic. 
Moreover we observe that the MLPL-HTSP heuristic provides remarkable improvements in terms of both worst case and best case, i.e. the maximum and minimum values of DEV in Table \ref{tab:compsol}. As far as the computing time is concerned, Table \ref{tab:comptime}  shows that MLPL-HTSP represents a good tradeoff  between  the baseline algorithm and the PL-HTSP. Indeed, the maximum computing time of MLPL-HTSP is remarkably lower than the minimum time of PL-HTSP, whilst the minimum computing time of MLPL-HTSP is only few seconds above the maximum time of HTSP. \\
\indent These results clearly illustrate that high quality results are obtained by the MLPL-HTSP algorithm for instances that correspond to realistic travel time functions.

\begin{table}[ht]
\scriptsize
\center
\caption{Impact of approximation $\underline{\tau}$ and the machine learning algorithm on solution quality}\label{tab:compsol}
\begin{tabular}{|c|c|c|c|c|}
\hline
\multicolumn{1}{|r|}{Testset} & Heuristic & \multicolumn{1}{r|}{Avg DEV\%} & \multicolumn{1}{r|}{min DEV} & \multicolumn{1}{r|}{max DEV} \\ \hline
London & HTSP      & 1.42\%  & 0.00   & 16.44 \\
London & PL-HTSP   & 0.35\%  & -0.90  & 8.36  \\
London & MLPL-HTSP & 0.23\%  & -0.49  & 8.16  \\ \hline
Paris  & HTSP      & 0.72\%  & -9.45  & 11.04 \\
Paris  & PL-HTSP   & -0.14\% & -13.13 & 13.13 \\
Paris  & MLPL-HTSP & -0.18\% & -12.52 & 3.93  \\ \hline
\end{tabular}
\end{table}

\begin{table}[ht]
\scriptsize
\center
\caption{Impact of approximation $\underline{\tau}$ and the machine learning algorithm on computing time}\label{tab:comptime}
\begin{tabular}{|c|c|c|c|c|}
\hline
\multicolumn{1}{|r|}{Testset} & Heuristic & \multicolumn{1}{r|}{Avg Time} & \multicolumn{1}{r|}{min Time} & \multicolumn{1}{r|}{max Time} \\ \hline
London & HTSP      & 1.26   & 0.08  & 7.18   \\
London & PL-HTSP   & 128.52 & 91.72 & 195.34 \\
London & MLPL-HTSP & 18.28  & 14.95 & 26.40  \\ \hline
Paris  & HTSP      & 1.94   & 0.06  & 10.90  \\
Paris  & PL-HTSP   & 83.12  & 57.11 & 105.93 \\
Paris  & MLPL-HTSP & 12.46  & 8.73  & 37.47  \\ \hline
\end{tabular}
\end{table}

\begin{table}[ht]
\scriptsize
\center
\caption{Computational results of MLPL-HTSP for the London testset}\label{results-london}
\begin{tabular}{|lcccc|l|ccccc|}
\cline{1-5} \cline{7-11}
Instance  & $BK$   & $\underline{z}_{\Omega}$ & $DEV\%$ & time  &  & Instance & $BK$   & $\underline{z}_{\Omega}$ & $DEV\%$ & time  \\ \cline{1-5} \cline{7-11} 
10\_I\_1  & 407.59 & 407.59 & 0.00\% & 17.90 &  & 10\_I\_6 & 399.36 & 399.36 & 0.00\% & 15.70 \\ \cline{1-5} \cline{7-11} 
10\_I\_10 & 379.27 & 387.43 & 2.15\% & 18.44 &  & 10\_I\_7 & 388.38 & 388.69 & 0.08\% & 19.75 \\ \cline{1-5} \cline{7-11} 
10\_I\_11 & 400.62 & 403.28 & 0.66\% & 21.28 &  & 10\_I\_9 & 369.03 & 369.79 & 0.21\% & 18.84 \\ \cline{1-5} \cline{7-11} 
10\_I\_12 & 401.17 & 402.09 & 0.23\% & 19.73 &  & 1\_I\_2  & 388.70 & 390.75 & 0.53\% & 18.53 \\ \cline{1-5} \cline{7-11} 
10\_I\_13 & 463.42 & 463.42 & 0.00\% & 24.86 &  & 1\_I\_26 & 419.04 & 419.04 & 0.00\% & 16.68 \\ \cline{1-5} \cline{7-11} 
10\_I\_14 & 399.75 & 399.77 & 0.01\% & 21.40 &  & 1\_I\_27 & 378.45 & 378.45 & 0.00\% & 16.43 \\ \cline{1-5} \cline{7-11} 
10\_I\_15 & 415.50 & 418.84 & 0.80\% & 18.34 &  & 1\_I\_28 & 393.14 & 394.52 & 0.35\% & 16.37 \\ \cline{1-5} \cline{7-11} 
10\_I\_16 & 401.62 & 401.81 & 0.05\% & 16.84 &  & 1\_I\_29 & 393.51 & 394.14 & 0.16\% & 23.73 \\ \cline{1-5} \cline{7-11} 
10\_I\_17 & 402.36 & 402.36 & 0.00\% & 15.60 &  & 1\_I\_3  & 396.82 & 399.36 & 0.64\% & 15.48 \\ \cline{1-5} \cline{7-11} 
10\_I\_19 & 436.13 & 436.13 & 0.00\% & 18.80 &  & 1\_I\_30 & 387.16 & 387.16 & 0.00\% & 15.33 \\ \cline{1-5} \cline{7-11} 
10\_I\_2  & 372.64 & \textbf{372.31}          & -0.09\% & 15.25 &  & 1\_I\_31 & 363.90 & 363.90                   & 0.00\%  & 14.95 \\ \cline{1-5} \cline{7-11} 
10\_I\_20 & 422.78 & 425.09 & 0.55\% & 17.53 &  & 1\_I\_32 & 408.21 & 408.21 & 0.00\% & 17.31 \\ \cline{1-5} \cline{7-11} 
10\_I\_23 & 400.79 & 400.82 & 0.01\% & 18.75 &  & 1\_I\_33 & 414.32 & 415.26 & 0.23\% & 21.69 \\ \cline{1-5} \cline{7-11} 
10\_I\_24 & 411.51 & 413.28 & 0.43\% & 18.93 &  & 1\_I\_34 & 365.65 & 365.94 & 0.08\% & 15.65 \\ \cline{1-5} \cline{7-11} 
10\_I\_25 & 404.39 & 404.64 & 0.06\% & 17.52 &  & 1\_I\_35 & 412.53 & 412.53 & 0.00\% & 19.08 \\ \cline{1-5} \cline{7-11} 
10\_I\_26 & 409.90 & 410.32 & 0.10\% & 18.72 &  & 1\_I\_36 & 369.79 & 374.14 & 1.18\% & 19.30 \\ \cline{1-5} \cline{7-11} 
10\_I\_27 & 420.02 & 420.02 & 0.00\% & 19.97 &  & 1\_I\_37 & 410.90 & 410.91 & 0.00\% & 16.71 \\ \cline{1-5} \cline{7-11} 
10\_I\_28 & 419.80 & 421.90 & 0.50\% & 19.94 &  & 1\_I\_39 & 406.39 & 407.94 & 0.38\% & 22.43 \\ \cline{1-5} \cline{7-11} 
10\_I\_29 & 408.59 & 409.82 & 0.30\% & 20.94 &  & 1\_I\_4  & 402.54 & 402.65 & 0.03\% & 26.40 \\ \cline{1-5} \cline{7-11} 
10\_I\_30 & 395.66 & 396.32 & 0.17\% & 15.70 &  & 1\_I\_40 & 396.62 & 396.62 & 0.00\% & 15.03 \\ \cline{1-5} \cline{7-11} 
10\_I\_31 & 409.23 & 411.73 & 0.61\% & 24.82 &  & 1\_I\_42 & 408.81 & 408.81 & 0.00\% & 20.21 \\ \cline{1-5} \cline{7-11} 
10\_I\_32 & 398.56 & \textbf{398.07}          & -0.12\% & 15.58 &  & 1\_I\_44 & 373.48 & 374.71                   & 0.33\%  & 21.97 \\ \cline{1-5} \cline{7-11} 
10\_I\_33 & 345.61 & 350.94 & 1.54\% & 17.12 &  & 1\_I\_45 & 367.21 & 367.26 & 0.01\% & 15.16 \\ \cline{1-5} \cline{7-11} 
10\_I\_34 & 353.48 & 353.52 & 0.01\% & 18.41 &  & 1\_I\_46 & 404.26 & 404.59 & 0.08\% & 17.55 \\ \cline{1-5} \cline{7-11} 
10\_I\_36 & 394.61 & 394.61 & 0.00\% & 15.91 &  & 1\_I\_47 & 402.02 & 402.61 & 0.15\% & 18.54 \\ \cline{1-5} \cline{7-11} 
10\_I\_37 & 416.03 & 416.59 & 0.13\% & 16.02 &  & 1\_I\_48 & 393.13 & 394.97 & 0.47\% & 16.31 \\ \cline{1-5} \cline{7-11} 
10\_I\_38 & 453.65 & 453.79 & 0.03\% & 19.90 &  & 1\_I\_49 & 381.64 & 381.64 & 0.00\% & 16.16 \\ \cline{1-5} \cline{7-11} 
10\_I\_39 & 426.38 & 426.49 & 0.03\% & 17.30 &  & 1\_I\_5  & 333.64 & 335.85 & 0.66\% & 15.96 \\ \cline{1-5} \cline{7-11} 
10\_I\_40 & 416.32 & 417.37 & 0.25\% & 18.13 &  & 1\_I\_50 & 372.23 & 372.62 & 0.10\% & 16.18 \\ \cline{1-5} \cline{7-11} 
10\_I\_41 & 398.48 & 398.48 & 0.00\% & 16.61 &  & 1\_I\_51 & 417.30 & 417.74 & 0.11\% & 18.47 \\ \cline{1-5} \cline{7-11} 
10\_I\_5  & 393.85 & 395.13 & 0.32\% & 19.25 &  & 1\_I\_53 & 405.22 & 405.22 & 0.00\% & 16.08 \\ \cline{1-5} \cline{7-11} 
\end{tabular}
\end{table}

\begin{table}[ht]
\scriptsize
\center
\caption{Computational results of MLPL-HTSP for the Paris testset}\label{results-paris}
\begin{tabular}{|lcccc|l|ccccc|}
\cline{1-5} \cline{7-11}
Instance & $BK$ & $\underline{z}_{\Omega}$ & $DEV\%$ & time &  & Instance & $BK$ & $\underline{z}_{\Omega}$ & $DEV\%$ & time \\ \cline{1-5} \cline{7-11} 
0\_I\_0   & 289.26 & 289.26          & 0.00\%  & 15.59 &  & 0\_I\_133 & 280.81 & 280.81          & 0.00\%  & 11.74 \\ \cline{1-5} \cline{7-11} 
0\_I\_1   & 282.15 & 282.23          & 0.03\%  & 11.54 &  & 0\_I\_134 & 287.89 & 288.53          & 0.22\%  & 37.47 \\ \cline{1-5} \cline{7-11} 
0\_I\_10  & 291.04 & 291.09          & 0.02\%  & 9.98  &  & 0\_I\_135 & 305.73 & \textbf{304.78} & -0.31\% & 11.78 \\ \cline{1-5} \cline{7-11} 
0\_I\_100 & 285.31 & 285.31          & 0.00\%  & 10.04 &  & 0\_I\_136 & 283.76 & \textbf{283.43} & -0.12\% & 10.25 \\ \cline{1-5} \cline{7-11} 
0\_I\_101 & 286.66 & \textbf{274.14} & -4.37\% & 18.64 &  & 0\_I\_137 & 279.85 & \textbf{279.47} & -0.14\% & 10.86 \\ \cline{1-5} \cline{7-11} 
0\_I\_102 & 273.71 & 273.88          & 0.06\%  & 9.55  &  & 0\_I\_138 & 275.06 & 275.06          & 0.00\%  & 9.10  \\ \cline{1-5} \cline{7-11} 
0\_I\_103 & 297.27 & 297.27          & 0.00\%  & 11.36 &  & 0\_I\_139 & 300.82 & \textbf{300.41} & -0.14\% & 11.04 \\ \cline{1-5} \cline{7-11} 
0\_I\_104 & 289.87 & 290.07          & 0.07\%  & 9.83  &  & 0\_I\_14  & 277.91 & \textbf{274.31} & -1.30\% & 9.31  \\ \cline{1-5} \cline{7-11} 
0\_I\_105 & 309.26 & 309.40          & 0.05\%  & 9.75  &  & 0\_I\_140 & 295.50 & \textbf{294.39} & -0.38\% & 10.12 \\ \cline{1-5} \cline{7-11} 
0\_I\_106 & 286.73 & 286.82          & 0.03\%  & 9.45  &  & 0\_I\_141 & 300.23 & \textbf{298.68} & -0.52\% & 13.27 \\ \cline{1-5} \cline{7-11} 
0\_I\_107 & 295.62 & 295.91          & 0.10\%  & 10.71 &  & 0\_I\_142 & 285.36 & \textbf{281.75} & -1.27\% & 11.91 \\ \cline{1-5} \cline{7-11} 
0\_I\_108 & 279.18 & \textbf{278.58} & -0.21\% & 9.67  &  & 0\_I\_143 & 287.65 & 287.65          & 0.00\%  & 10.39 \\ \cline{1-5} \cline{7-11} 
0\_I\_109 & 287.85 & 287.85          & 0.00\%  & 15.48 &  & 0\_I\_144 & 277.19 & \textbf{276.35} & -0.30\% & 9.32  \\ \cline{1-5} \cline{7-11} 
0\_I\_11  & 310.77 & 310.77          & 0.00\%  & 11.61 &  & 0\_I\_145 & 254.78 & 255.08          & 0.12\%  & 8.79  \\ \cline{1-5} \cline{7-11} 
0\_I\_110 & 274.52 & 278.46          & 1.44\%  & 10.69 &  & 0\_I\_146 & 288.52 & 288.62          & 0.03\%  & 12.47 \\ \cline{1-5} \cline{7-11} 
0\_I\_111 & 301.50 & \textbf{300.51} & -0.33\% & 14.62 &  & 0\_I\_147 & 295.02 & \textbf{292.48} & -0.86\% & 11.59 \\ \cline{1-5} \cline{7-11} 
0\_I\_112 & 306.67 & \textbf{305.80} & -0.28\% & 15.94 &  & 0\_I\_148 & 276.02 & 276.24          & 0.08\%  & 8.73  \\ \cline{1-5} \cline{7-11} 
0\_I\_113 & 303.81 & 306.41          & 0.86\%  & 14.32 &  & 0\_I\_149 & 289.43 & 289.69          & 0.09\%  & 9.88  \\ \cline{1-5} \cline{7-11} 
0\_I\_114 & 298.17 & \textbf{296.57} & -0.54\% & 14.69 &  & 0\_I\_15  & 299.90 & 299.90          & 0.00\%  & 10.92 \\ \cline{1-5} \cline{7-11} 
0\_I\_115 & 293.19 & 294.04          & 0.29\%  & 10.71 &  & 0\_I\_150 & 290.86 & \textbf{289.40} & -0.50\% & 11.92 \\ \cline{1-5} \cline{7-11} 
0\_I\_116 & 288.90 & 288.90          & 0.00\%  & 24.52 &  & 0\_I\_151 & 283.60 & 283.77          & 0.06\%  & 11.90 \\ \cline{1-5} \cline{7-11} 
0\_I\_117 & 300.82 & \textbf{297.73} & -1.03\% & 10.92 &  & 0\_I\_152 & 293.53 & \textbf{287.85} & -1.94\% & 10.88 \\ \cline{1-5} \cline{7-11} 
0\_I\_118 & 275.94 & 275.98          & 0.01\%  & 10.36 &  & 0\_I\_153 & 273.22 & 273.22          & 0.00\%  & 10.88 \\ \cline{1-5} \cline{7-11} 
0\_I\_119 & 274.69 & 274.69          & 0.00\%  & 9.65  &  & 0\_I\_154 & 289.59 & \textbf{288.51} & -0.37\% & 10.20 \\ \cline{1-5} \cline{7-11} 
0\_I\_12  & 301.23 & 302.65          & 0.47\%  & 12.61 &  & 0\_I\_155 & 318.15 & 318.15          & 0.00\%  & 10.41 \\ \cline{1-5} \cline{7-11} 
0\_I\_120 & 295.00 & 295.08          & 0.03\%  & 11.40 &  & 0\_I\_156 & 278.43 & 278.69          & 0.09\%  & 9.34  \\ \cline{1-5} \cline{7-11} 
0\_I\_121 & 289.19 & 289.31          & 0.04\%  & 10.39 &  & 0\_I\_157 & 292.37 & \textbf{288.54} & -1.31\% & 11.95 \\ \cline{1-5} \cline{7-11} 
0\_I\_122 & 283.25 & \textbf{281.89} & -0.48\% & 17.13 &  & 0\_I\_159 & 292.76 & 294.04          & 0.44\%  & 12.60 \\ \cline{1-5} \cline{7-11} 
0\_I\_123 & 312.11 & 312.12          & 0.00\%  & 11.90 &  & 0\_I\_16  & 304.56 & \textbf{301.95} & -0.86\% & 20.07 \\ \cline{1-5} \cline{7-11} 
0\_I\_124 & 300.24 & \textbf{298.42} & -0.61\% & 14.63 &  & 0\_I\_160 & 281.17 & 281.40          & 0.08\%  & 12.63 \\ \cline{1-5} \cline{7-11} 
0\_I\_125 & 285.50 & 285.64          & 0.05\%  & 9.42  &  & 0\_I\_161 & 305.14 & 305.14          & 0.00\%  & 11.13 \\ \cline{1-5} \cline{7-11} 
0\_I\_126 & 296.42 & 297.22          & 0.27\%  & 21.34 &  & 0\_I\_162 & 335.01 & \textbf{334.37} & -0.19\% & 10.93 \\ \cline{1-5} \cline{7-11} 
0\_I\_127 & 299.22 & 299.25          & 0.01\%  & 10.28 &  & 0\_I\_163 & 289.14 & \textbf{287.52} & -0.56\% & 15.06 \\ \cline{1-5} \cline{7-11} 
0\_I\_128 & 285.49 & 285.64          & 0.05\%  & 15.82 &  & 0\_I\_164 & 272.99 & \textbf{272.87} & -0.04\% & 10.77 \\ \cline{1-5} \cline{7-11} 
0\_I\_129 & 282.04 & 282.04          & 0.00\%  & 9.48  &  & 0\_I\_165 & 290.55 & 290.73          & 0.06\%  & 10.09 \\ \cline{1-5} \cline{7-11} 
0\_I\_13  & 287.11 & 287.11          & 0.00\%  & 13.10 &  & 0\_I\_166 & 308.36 & 308.57          & 0.07\%  & 12.85 \\ \cline{1-5} \cline{7-11} 
0\_I\_130 & 315.47 & \textbf{314.00} & -0.47\% & 15.41 &  & 0\_I\_168 & 304.05 & 304.05          & 0.00\%  & 9.78  \\ \cline{1-5} \cline{7-11} 
0\_I\_131 & 271.56 & 271.57          & 0.00\%  & 9.94  &  & 0\_I\_169 & 280.77 & 280.90          & 0.05\%  & 9.83  \\ \cline{1-5} \cline{7-11} 
0\_I\_132 & 259.81 & \textbf{259.75} & -0.02\% & 14.29 &  & 0\_I\_17  & 309.58 & \textbf{309.05} & -0.17\% & 22.79 \\ \cline{1-5} \cline{7-11} 
\end{tabular}
\end{table}

\section{Conclusions}
The main contribution of this paper is an algorithm that learns from past data to solve the TDTSP in an efficient and effective manner.
Computational results on two European cities show that the average gap with the best-known solutions is only 0.001\% and the average computation time is 15 seconds. Furthermore, new best solutions have been produced for several test instances. This is achieved by solving a time-invariant Asymmetric TSP, where the arc (constant) costs are suitably defined by the combined use of an LP-based approach and a mix of unsupervised and supervised Machine Learning techniques. In particular, we make use of the ETA predictions provided by a feedforward neural network trained on past instances solved to optimality or near-optimality.
With regard to future research we want to investigate the definition of new features for the neural network as well as to exploit the use of deep learning methods \cite{goodfellow2016deep}. Another noteworthy research goal concerns the study of a more efficient algorithm for (approximately) minimizing the fitting deviation between the travel time function $\tau$ and its \textit{approximation} $\underline{\tau}_{\Omega}$. Finally, future research could be focused on the adaptation of the ideas introduced in this paper to other routing problems.

 \section*{Conflict of interest}
 The authors declare that they have no conflict of interest.


\bibliographystyle{spmpsci}      
\bibliography{biblio.bib}

\end{document}